\documentclass[10pt,twocolumn,letterpaper]{article}

\usepackage{cvpr}              

\usepackage{graphicx}
\usepackage{amsmath}
\usepackage{amssymb}
\usepackage{booktabs}
\usepackage{colortbl}
\usepackage{float}
\usepackage[table,dvipsnames]{xcolor}
\usepackage{multirow}
\usepackage{amssymb}
\usepackage{pifont}
\newcommand{\xmark}{\ding{55}}%

\usepackage[accsupp]{axessibility}
\usepackage[pagebackref,breaklinks,colorlinks]{hyperref}

\usepackage[capitalize]{cleveref}
\crefname{section}{Sec.~}{Sec.~s.}
\Crefname{section}{Sec.~s}{Sec.~s}
\Crefname{table}{Tab.~s}{Table.s}
\crefname{table}{Tab.~}{Tabs.}

\def\cvprPaperID{6951} 
\def\confName{CVPR}
\def\confYear{2023}

\begin{document}

\title {Real-time Controllable Denoising for Image and Video }


\author{\normalsize
Zhaoyang Zhang\textsuperscript{\scriptsize{1}}\thanks{Equal contribution.}\quad Yitong Jiang\textsuperscript{\scriptsize{1}}\footnotemark[1]\quad Wenqi Shao\textsuperscript{\scriptsize{1,3}}\quad  Xiaogang Wang\textsuperscript{\scriptsize{1}}\quad Ping Luo\textsuperscript{\scriptsize{2,3}}\quad Kaimo Lin\textsuperscript{4}\quad Jinwei Gu\textsuperscript{\scriptsize{1,3} } \\\
{\small \textsuperscript{\scriptsize{1}}The Chinese University of Hong Kong
 \qquad \textsuperscript{\scriptsize{2}}The Univesity of Hong Kong
 \qquad \textsuperscript{\scriptsize{3}}Shanghai AI Laboratory
 \qquad \textsuperscript{\scriptsize{4}}SenseBrain} \\
{\tt\small \{zhaoyangzhang@link, ytjiang@link,weqish@link, xgwang@ee,jwgu@\}.cuhk.edu.hk}  \\
{\tt\small linkaimo1990@gmail.com,  pluo@cs.hku.hk}
}

\maketitle







%


\begin{abstract}
Controllable image denoising aims to generate clean samples with human perceptual priors and balance sharpness and smoothness. In traditional filter-based denoising methods, this can be easily achieved by adjusting the filtering strength. However, for NN  (Neural Network)-based models, adjusting the final denoising strength requires performing network inference each time, making it almost impossible for real-time user interaction.
In this paper, we introduce Real-time Controllable Denoising  (RCD), the first deep image and video denoising pipeline that provides a fully controllable user interface to edit arbitrary denoising levels in real-time with only one-time network inference. Unlike existing controllable denoising methods that require multiple denoisers and training stages, RCD replaces the last output layer  (which usually outputs a single noise map) of an existing CNN-based model with a lightweight module that outputs multiple noise maps. We propose a novel Noise Decorrelation process to enforce the orthogonality of the noise feature maps, allowing arbitrary noise level control through noise map interpolation. This process is network-free and does not require network inference. Our experiments show that RCD can enable real-time editable image and video denoising for various existing heavy-weight models without sacrificing their original performance.

\end{abstract}

\section{Introduction}




Image and video denoising are fundamental problems in computational photography and computer vision. With the development of deep neural networks~\cite{he2016deep,creswell2018generative,vaswani2017attention,zhang2021star}, model-based denoising methods  have achieved tremendous success in generating clean images and videos with superior denoising scores~\cite{zhang2017beyond,zhang2018ffdnet,anwar2019real}.
However, it should be noted that the improvement in reconstruction accuracy  (e.g., PSNR, SSIM) is not always accompanied by an improvement in visual quality, which is known as the Perception-Distortion trade-off~\cite{blau2018perception}.
In traditional denoising approaches, we can easily adjust the denoising level by tuning related control parameters and deriving our preferred visual results. However, for typical deep network methods, we can only restore the degraded image or video to a fixed output with a predetermined restoration level.


\begin{figure}
\includegraphics[width=1\linewidth]{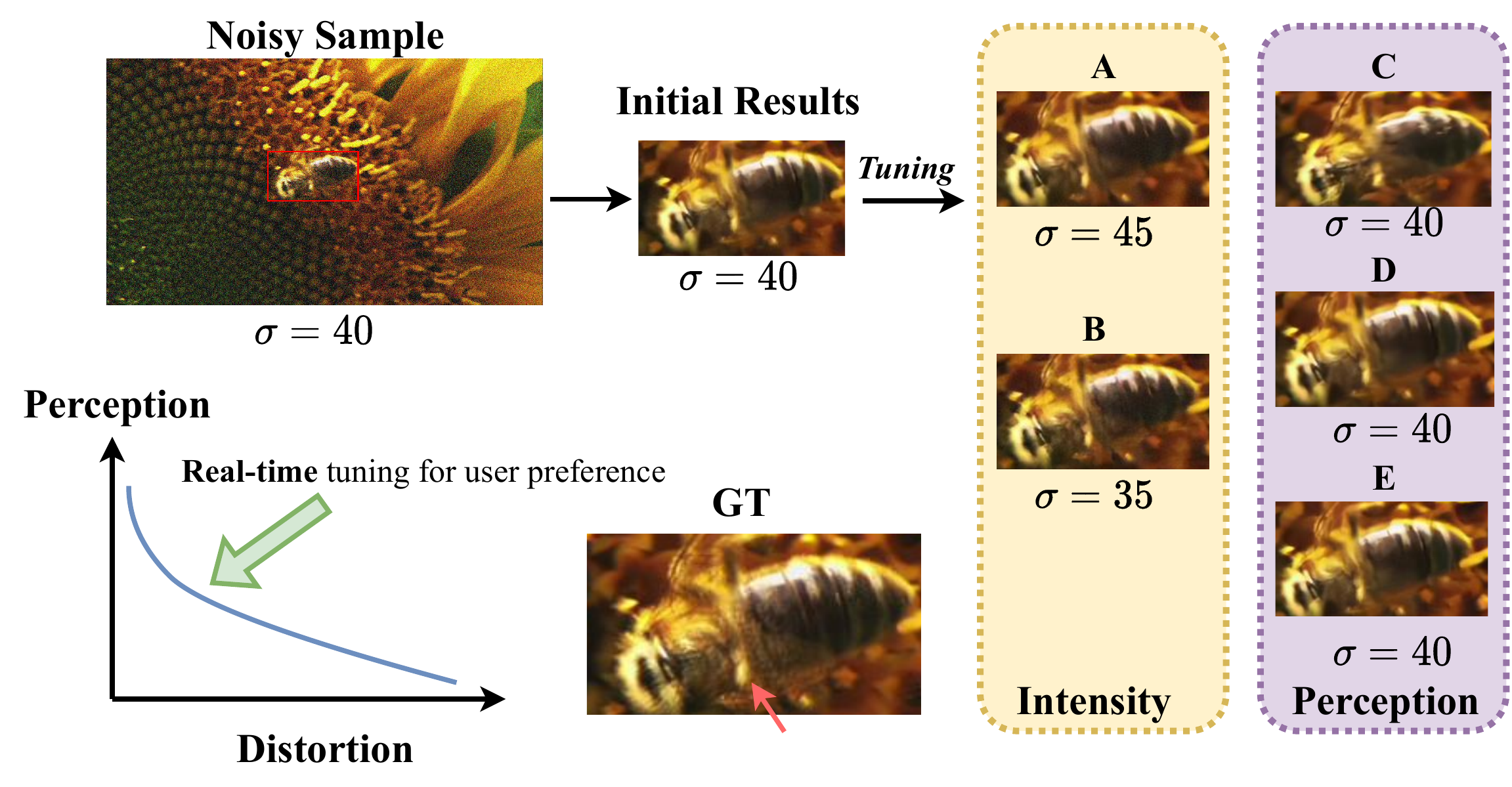}
\centering
\caption{ \small Real-time controllable denoising allows users further tuning the restored results to achieve Perception-Distortion trade-off. \textbf{A-B}: tuning with changing denoising intensity. \textbf{C-E}: tuning without changing denoising intensity. }
\label{fig:teaser}
\vspace{-15pt}
\end{figure}

In recent years, several modulation methods have been proposed to generate continuous restoration effects between two pre-defined denoising levels. These methods can be categorized into two kinds: interpolation-based methods~\cite{wang2019deep,fan2018decouple,he2019modulating,wang2019cfsnet}, which use deep feature interpolation layers, and condition-network-based methods, which import an extra condition network for denoising control~\cite{he2020interactive,cai2021toward,mou2022metric}. Essentially, both types of methods are designed based on the observation that the outputs of the network change continuously with the modulation of features/filters. This observation enables deep denoising control, but it also introduces several limitations.
First, there is a \textbf{lack of explainability}, as the relationship between the control parameters  (how to modulate features) and the control operation  (how the network outputs are changed) is unclear~\cite{he2019modulating}. This indicates that black-box operators  (network layers) must be used to encode them. Second, the use of control parameters as network inputs requires entire network propagation each time control parameters change, resulting in a \textbf{lack of efficiency}. Lastly, current modulation methods often require an explicit degradation level during training, which is hard to obtain for \textbf{real-world samples}. As a result, current controllable denoising methods only focus on synthetic noise benchmarks.
Furthermore, both interpolation-based and condition-network-based methods have their own drawbacks. Interpolation-based methods often require multiple training stages, including pretraining two basic models  (start level and end level). On the other hand, condition-network-based methods are strenuous to jointly optimize the base network and the condition network.

\begin{figure}
\includegraphics[width=1\linewidth]{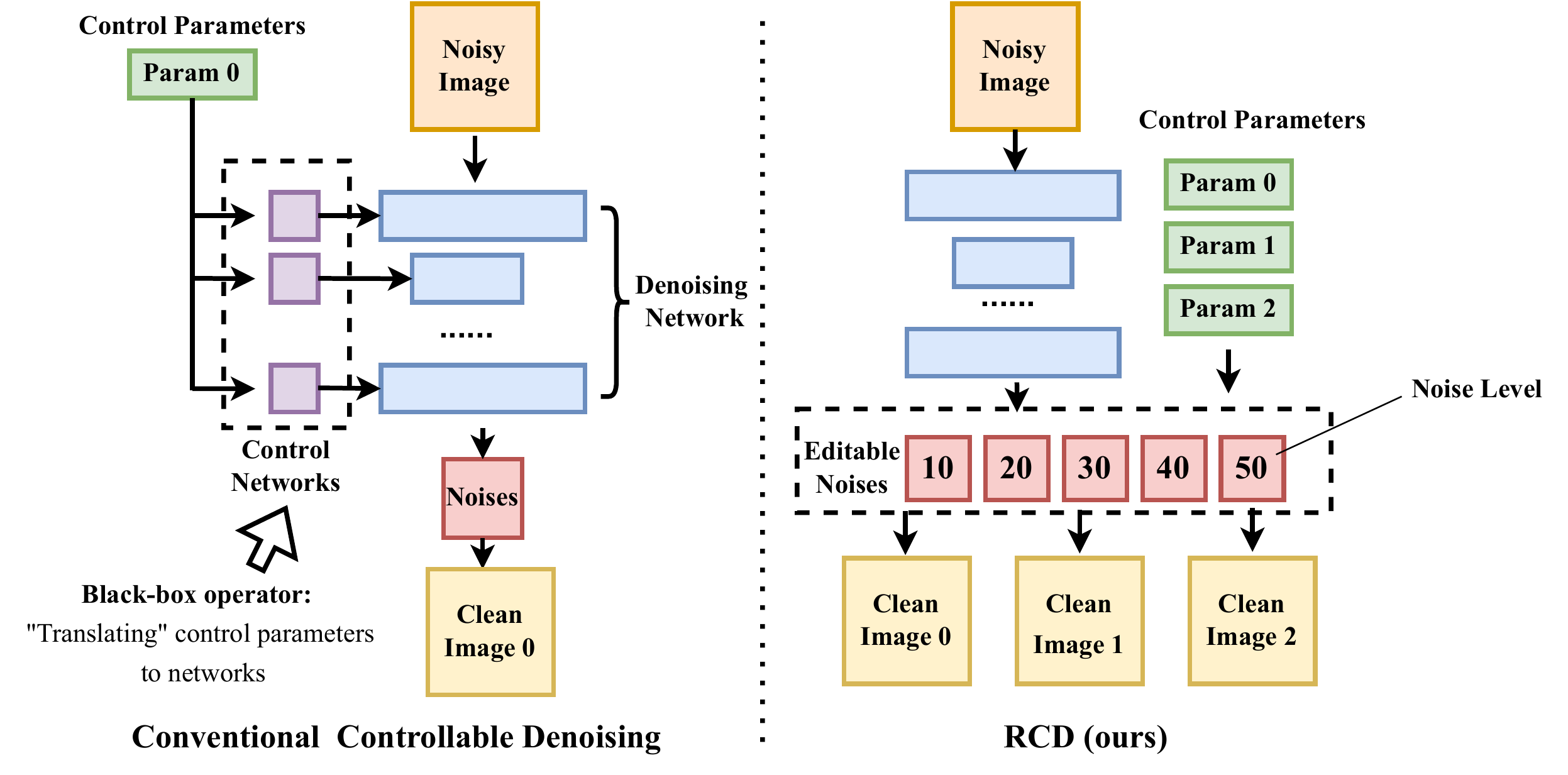}
\centering
\caption{\small Comparison of pipelines between conventional controllable denoising and our RCD. RCD achieves real-time noise control by manipulating editable noises directly.}
\label{fig:sum}
\vspace{-10pt}
\end{figure}

In this paper, we research on the problem:  Can we achieve real-time controllable denoising that abandons the auxiliary network and requires no network forward propagation for changing restoration effects at test time?

Towards this goal, we propose Real-time Controllable Denoising method  (RCD), a lightweight pipeline for enabling rapid denoising control to achieve Perception-Distortion Balance  (See Fig.~\ref{fig:teaser}). 
Our RCD can be plugged into any noise-generate-based restoration methods~\cite{chen2022simple,zamir2022restormer,tassano2020fastdvdnet,zhang2017beyond} with just a few additional calculations. 
Specifically, we replace the last layer of an existing denoising network  (which usually outputs a single noise map) with a lightweight module that generates multiple noise maps with different noise levels. We utilize a novel Noise Decorrelation process to enforce the orthogonality of the noise distribution of these noise maps during training. As a result, we can attain arbitrary denoising effects by simple linear interpolation of these noise maps. Since this process does not require network inference, it makes real-time user interaction possible even for heavy denoising networks.

Fig.~\ref{fig:sum} illustrates the fundamental differences between our RCD approach and conventional controllable denoising methods. In contrast to traditional methods that rely on control networks, the RCD pipeline generates editable noises of varying intensities/levels, providing explicit control by external parameters and enabling network-free, real-time denoising editing.
Real-time editing capabilities offered by RCD create new opportunities for numerous applications that were previously impossible using conventional techniques, such as online video denoising editing, even during playback  (e.g., mobile phone camera video quality tuning for ISP tuning engineers), as well as deploying controllable denoising on edge devices and embedded systems. Since the editing stage of RCD only involves image interpolation, users can edit their desired results on low-performance devices without the need for GPUs/DSPs.

Moreover, unlike previous methods that only support changing noise levels, RCD allows users to adjust denoising results at a specific noise level by providing a new interface to modify the noise generation strategy. RCD is also the first validated method for controllable denoising on real-world benchmarks. It is noteworthy that existing controllable methods typically require training data with fixed-level noise to establish their maximum and minimum noise levels, which makes them unsuitable for most real-world benchmarks comprising data with varying and unbalanced noise levels.



%
Our main contributions can be summarized as follows:
\begin{itemize}
  \setlength\itemsep{-0.2em}
  \item We propose RCD, a controllable denoising pipeline that firstly supports \textbf{real-time denoising control}  ($>\textbf{2000}\times$ speedup compared to conventional controllable methods) and  \textbf{larger control capacity}  (more than just intensity) without  multiple training stages~\cite{he2019modulating} and  auxiliary networks~\cite{wang2019cfsnet}.
  \item RCD is the first method supporting controllable denoising on real-world benchmarks.
  \item We propose a general Noise Decorrelation technique to estimate editable noises.
  \item We achieve comparable or better results on widely-used real/synthetic image-denoising and video-denoising datasets with minimal additional computational cost.
  

\end{itemize}



%
%
%


\section{Related Works}
\subsection{Denoising}
Traditional image and video denoising methods are often based on prior assumptions such as sparse image prior \cite{aharon2006k,elad2006image,getreuer2018blade,dong2011sparsity}, non-local similarity \cite{buades2005non,foi2007pointwise,dabov2008image,dabov2007image}, and other similar techniques \cite{gu2014weighted,portilla2003image,xu2007iterative}. However, with the recent development of deep learning networks, many learning-based methods have been proposed and achieved state-of-the-art performance. Early works \cite{burger2012image} utilized multi-layer perceptron  (MLP) to achieve comparable results with BM3D. In recent years, there has been rapid progress on CNN-based denoising methods \cite{zhang2017beyond,zhang2018ffdnet,tian2020image,gu2019self,anwar2019real,cai2021learning} and Transformer-based methods \cite{zamir2022restormer,liang2022vrt,zhang2021star,shao2021dynamic}, which have started to dominate the image/video denoising task.
However, the above-mentioned works mainly focus on designing novel network architectures to improve the denoising performance and usually generate a single output. Their lack of ability to adjust the output denoising level based on user's feedback has greatly restricted their practical use in many real-world applications. Moreover, although techniques like pruning \cite{molchanov2019importance,zhang2019differentiable,liu2018rethinking} and quantization \cite{tao2021fat, zhang2021differentiable} can accelerate such neural network-based methods, they are typically heavy, which restricts their application to real-time denoising control.

%

\begin{figure*}
\vspace{-10pt}
\includegraphics[width=1\linewidth]{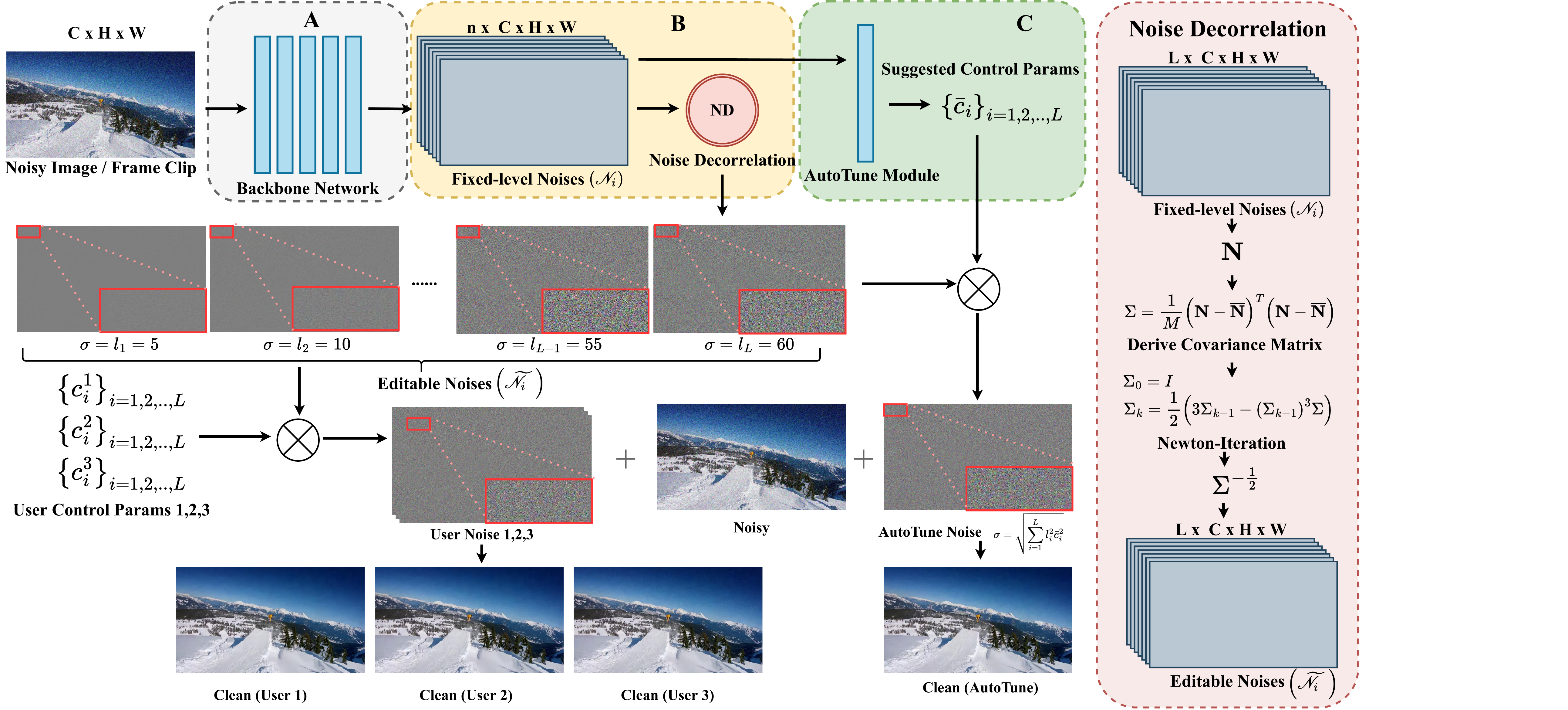}
\centering
\vspace{-15pt}

\caption{\small Pipeline overview of proposed RCD framework. \textbf{A:} Backbone network for  generating multi-level noise maps. \textbf{B:} Noise Decorrelation module for editable noises. \textbf{C: } AutoTune module for providing reference control parameters for users.  }
\label{fig:pipeline}
\vspace{-10pt}

\end{figure*}

\subsection{Controllable denoising}
Most conventional deep-learning methods for image/video denoising can only generate a fixed result with a specific restoration level.
Recently, some controllable image/video denoising methods allow users to adjust the restoration effect without retraining the network. 
DNI~\cite{wang2019deep} and AdaFM~\cite{he2019modulating} used the observation that the learned filters of the models trained with different restoration levels are similar in visual patterns. 
DNI interpolated all the corresponding parameters between two related networks to derive smooth and continuous restoration effects, while AdaFM adopted feature modulation filters after each convolution layer. 
CFSNet~\cite{wang2019cfsnet} proposed an adaptive learning strategy of using interpolation coefficients to couple the intermediate features between the main branch and the tuning branch. 
Different from these interpolation-base methods, some other methods~\cite{he2020interactive,cai2021toward,mou2022metric} regarded modulation as a conditional image restoration problem and adopted a joint training strategy. 
CUGAN~\cite{cai2021toward} proposed a GAN-based image restoration framework to avoid the over-smooth problem, a common issue in PSNR-oriented methods.
However, all of the above controllable methods can only be trained with synthetic degradations because they require explicit degradation levels during training.
When applied on real-world data, as shown in~\cite{guo2019toward}, methods that trained for blind
Additive White Gaussian Noise  (AWGN)~\cite{zhang2017beyond, mao2016image} may be overfitted and often suffer from dramatic performance drop.
Besides the real-world image issue, all these controllable methods utilize an auxiliary conditional network and require one network inference for each different target restoration level at test time, which makes them almost impossible for real-time application.

\section{Methods}

\subsection{Conventional Deep Denoising}
Deep denoising methods trump traditional filter-based techniques by leveraging the neural networks' robust representation learning capability. 
Most current denoising methods~\cite{chen2022simple,liang2022vrt,tassano2020fastdvdnet} reason about the relationship between clean and noisy images by regressing noise maps with a neural generator. 
Specifically, given a noisy image $\mathbf{I_n}$ and model $\mathcal{M}: \mathbb{R}^{H\times W \times C} \to \mathbb{R}^{H\times W \times C} $, we can derive the predicted clean image $\mathbf{I_c}$ by: 
    $\mathbf{I_c} = \mathbf{I_n} + \mathcal{M} (\mathbf{I_n})$,
where model $\mathcal{M}$ is updated by minimizing the distance between the denoising result $\mathbf{I_c}$ and the ground truth $\mathbf{I_{gt}}$. As we can see, this kind of approach generates a single fixed output result in a black-box manner, making it almost impossible to adjust the denoise operation explicitly. 




\subsection{Pipeline Overview}

In this section, we present Real-time Controllable Denoising  (RCD), a novel deep learning-based pipeline for real-time controllable denoising.
As illustrated in Fig.~\ref{fig:pipeline}, RCD essentially consists of three parts:  (1) A backbone network, \ie, $\mathcal{M}_{b} : \mathbb{R}^{H\times W \times C} \to \mathbb{R}^{H\times W \times LC}$, generates multiple fixed-level noise maps, where $L$ is the number of pre-defined noise levels  (see  (A) in Fig.~\ref{fig:pipeline}).  (2) A Noise Decorrelation (ND) block that enforces the editability of the generated noise maps  (see  (B) in Fig.~\ref{fig:pipeline}).  (3) An AutoTune module that gives a default set of control parameters to generate the best denoising result.

Specifically, the backbone network will generate multiple fixed-level noise maps, \ie, $\{{\mathcal{N}}_i\}_{i=1}^L$, for each noisy image input. The noise maps are then fed into the proposed Noise Decorrelation (ND) block, which makes noise maps orthogonal to each other. In this way, the decorrelated noise maps $\{\Tilde{\mathcal{N}}_i\}_{i=1}^L$ will be zero-correlated and thus become linearly interpolable. At last, the AutoTune module will give a set of suggested control parameters $\{ \bar{c}_i \}_{i=1}^L$ to generate the final denoising result as follows:
\begin{equation}\label{eq:overall_RCD}
     \mathbf{I_c} = \mathbf{I_n} + \sum_{i=1}^{L}\bar{c}_i \tilde{\mathcal{N}}_i, 
\end{equation}
where $\sum_{i=1}^L \bar{c}_i = 1$.
Moreover, given the zero-correlated noise maps, users can also generate arbitrary strength denoising results by replacing $\{ \bar{c}_i \}_{i=1}^L$ with their own customized control parameters $\{ c_i \}_{i=1}^L$.




\subsection{ Multi-level Noise Generation}
Given a noisy input image $\mathbf{I_n}$, the backbone network aims to  generate multiple noise maps $\{\mathcal{N}_i\}_{i=1}^L$, corresponding to a set of pre-defined noise levels $\{l_i\}_{i=1}^L$, \eg noise levels $\{5, 10, 15, ..., 60\}$. Hence, we have
\begin{equation}
    \sigma (\mathcal{N}_{i}) = l_i, \forall i = 1, ..., L,
\end{equation}
where $\sigma$ is the noise level operation that calculates the standard deviation of pixels in each noise map.
To obtain multi-level noise maps, we replace the conventional last output layer of the denoising network with a convolutional layer with an output channel size of $L\cdot C$. Moreover, the level of the noise map is explicitly generated with the normalization operation, as given by
\begin{equation}\label{eq:level}
    \mathcal{N}_{i} =  l_i \frac{\mathcal{M}_{b} (\mathbf{I_n})^{ (i)}}{ \sigma (\mathcal{M}_{b} (\mathbf{I_n})^{ (i)}) }, \forall i = 1, ..., L.
\end{equation}
Here $\mathcal{M}_{b} (\mathbf{I_n}) \in \mathbb{R}^{H\times W \times LC}$ is network output, and $\mathcal{M}_{b} (\mathbf{I_n})^{ (i)} \in \mathbb{R}^{H\times W \times C}$ is the i-th component separated from the channel dimension. The derived $\mathcal{N}_{i}$ can be considered as the noise map estimated at the given noise level $l_i$.

Different from prior controllable denoising methods with implicit interpolation in the network, we propose to explicitly interpolate the noise maps in Eqn.~\ref{eq:level}. Thanks to the separation of noise interpolation and network inference, our RCD can achieve real-time user interaction.

However, the multi-level noise maps $\mathcal{N}_{i}$ directly obtained by convolutional layers are usually highly correlated, which leads to the problem of noise level collapse. In other words, the noise map representations in different levels are redundant, implying that the number of noise maps at different noise strengths that participate in the linear interpolation in Eqn.~\ref{eq:overall_RCD} is implicitly reduced. 
Without any constraint, our experiments show that the single noise map at a certain noise level would dominate in the linear interpolation for a variety of input noisy images.
To address this issue, we further introduce the Noise Decorrelation block to make representations of these noise maps much more informative in the following section. 


\begin{figure}
\includegraphics[width=1\linewidth]{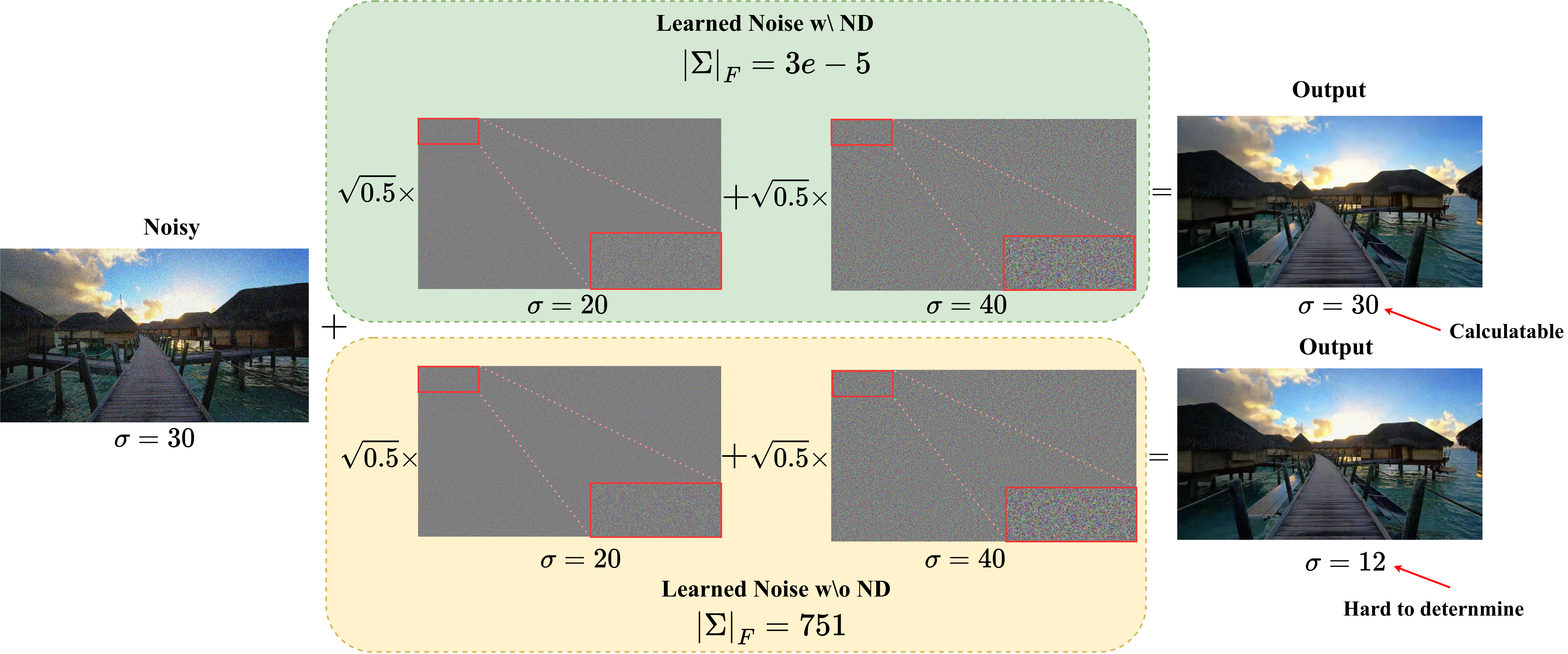}
\centering
\caption{\small Demonstration of Noise Decorrelation's influence on noise editing.  $|\Sigma|_F$ denotes norms of the covariance matrix for corresponding learned noises and $\sigma$ is noise intensity. }
\label{fig:ND}
\vspace{-10pt}
\end{figure}
\subsection{Noise Decorrelation}
The Noise Decorrelation (ND) block is designed to regularize the backbone network to generate editable noise maps at varying levels.
In particular, this block is a parameter-free computational unit that enforces $\mathcal{N}_{i}$ to be approximately zero-correlated with each other: $\mathbf{cov} (\mathcal{N}_{i}, \mathcal{N}_{j}) \approx 0, \forall i, j \in \{1, 2, ..., L\}$, where $\mathbf{cov} (\cdot,\cdot)$ is covariance operator. 
Inspired by the success of using the decorrelation technique in network optimization and normalization, we adopt the whitening-based methods~\cite{huang2018decorrelated, huang2019iterative} for noise decorrelation here.
The noise maps are decorrelated using the inverse square root of their covariance matrix. 

Specifically, for each predicted fixed-level noise map $\mathcal{N}_{i}$, it will firstly be reshaped to $\mathcal{N}_{i} \in \mathbb{R}^{1 \times M}$, where $M=HWC$. By stacking the reshaped $\mathcal{N}_{i}$ over the first dimension, we have $\mathbf{N}\in \mathbb{R}^{L \times M} $. We then calculate the noise covariance matrix $\sigma$ by:
    $\Sigma = \frac{1}{M-1}  (\mathbf{N} - \bar{\mathbf{N} })  (\mathbf{N} - \bar{\mathbf{N} }) ^T$
where $\bar{\mathbf{N}}$ is mean of $\mathbf{N}$ over channel $M$.

The Noise Decorrelation block needs to compute inverse square root $\Sigma^{-\frac{1}{2}} \in \mathbb{R}^{L\times L}$, which can be done by eigen decomposition or SVD. Since this kind of operation involves heavy computation~\cite{huang2018decorrelated}, we instead adopt the more
efficient Newton’s Iteration to estimate $\Sigma^{-\frac{1}{2}}$ as in~\cite{bini2005algorithms, higham1986newton}. Giving a covariance matrix $\Sigma$, Newton’s
Iteration calculates $\Sigma^{-\frac{1}{2}}$ by following the iterations below:
\begin{equation}
  \begin{array}{l}
    \Sigma_0 = I, \\
    \Sigma_k = \frac{1}{2}  (3\Sigma_{k-1} -  (\Sigma_{k-1})^3\Sigma), k=1,2,..,T,
  \end{array}
\end{equation}
where $k$ is the iteration index and $T$ is the iteration number  (in our experiments $T$ = 3 or 4). $\Sigma_k$ is guaranteed to converge to $\Sigma^{-\frac{1}{2}}$, if $\left\| I - \Sigma \right\|_{2} <  1$~\cite{bini2005algorithms}. This condition can be achieved by normalizing $\Sigma$ to $\frac{\Sigma}{tr (\Sigma)}$, where $tr (.)$ is trace operator. 

The derived $\Sigma^{-\frac{1}{2}}$ can be regarded as a whitening matrix~\cite{shao2021bwcp}, which decorrelates the noise maps $\mathbf{N}$ in a differentiable manner. The decorrelated noise maps $\Tilde{\mathbf{N} \in \mathbb{R}^{H\times W \times LC} }$ can be obtained by calculating:
    $\Tilde{\mathbf{N}} = \Sigma^{-\frac{1}{2}}\mathbf{N}$.
We can then have our editable fixed-level noises $\Tilde{\mathcal{N}}_i \in \mathbb{R}^{H\times W \times C} $ by reshaping $\Tilde{\mathbf{N}} $ and splitting it into $L$ noise maps. After Noise Decorrelation, we apply the same normalization as Eqn.~\ref{eq:level} to guarantee the noise strength of the decorrelated noises.

The zero-correlated noise maps $\Tilde{\mathcal{N}}_i$ present several excellent properties for controllable denoising. Firstly, the linearity of the noise level's square towards  $\Tilde{\mathcal{N}}_i$ is guaranteed. In other words, given an arbitrary set of control parameters $\{ c_i \}_{i=1}^L$, we  have 
\begin{equation} \label{eq:var}
   \mathbf{Var}  (\sum_{i=1}^{L} c_i \Tilde{\mathcal{N}}_i ) = \sum_{i=1}^{L} c_i^2 \mathbf{Var} (\Tilde{\mathcal{N}}_i),
\end{equation}
where $\mathbf{Var} (\cdot)$ denotes variance operator. Apparently, Eqn.~\ref{eq:var} holds when elements of $\{\Tilde{\mathcal{N}}_i\}_{i=1,2,..,m}$ are mutually zero-correlated. 
Eqn.~\ref{eq:var} reveals the explicit relationship between the control parameters and the target noise level, which allows us to directly edit noises by interpolating $\mathcal{N}_i$ using $c_i$.
%
Secondly, the Noise Decorrelation block can be regarded as a regularization tool that forces models to learn different noise formats for each level, which will increase the representation capacity of the denoising network~\cite{huang2019iterative}. 

Fig.~\ref{fig:ND} demonstrates how the Noise Decorrelation block works. 
With ND block, the covariance of learned noises is reduced to almost zero  (without it, $|\Sigma|_F$ can be 751 and unignorable), allowing us to derive determined interpolated results with target noise intensity. 
In contrast, without the Noise Decorrelation blocks, the output noise level can not be guaranteed.

\label{sec:3.4}

\subsection{AutoTune Module}
Given the decorrelated noise maps from the Noise Decorrelation block, the AutoTune module will predict a set of model-suggested control parameters, \ie, $\{ \bar{c}_i \}_{i=1}^L$, to generate the default denoising result. Users can then use this set of parameters as a starting point to fine-tune their final desired denoising strength. Our AutoTune module is extremely lightweight, and is formulated as a single-layer module with temperature softmax activation. 
Specifically, $\{ \bar{c}_i \}_{i=1}^L$ can be obtained by  :
$\bar{c}_i = \frac{e^{   \frac{\mathcal{A} (f)_i} {\tau }   }}{ \sum_{j=1}^n e^{   \frac{\mathcal{A} (f)_j} {\tau}   }   },$
where $\mathcal{A}$ is the NN layer, $f$ is the input feature maps, and $\tau$ is temperature. In our experiments, $\tau$ is set to be $0.05$ for best performance.
Following the design ethos of efficiency and least coupling to the backbone architecture, we directly choose the unnormalized model outputs $\mathcal{M} (\mathbf{I_n})$ as $f$  (see  (C) in Fig.~\ref{fig:pipeline}).

\begin{figure}
\includegraphics[width=1\linewidth]{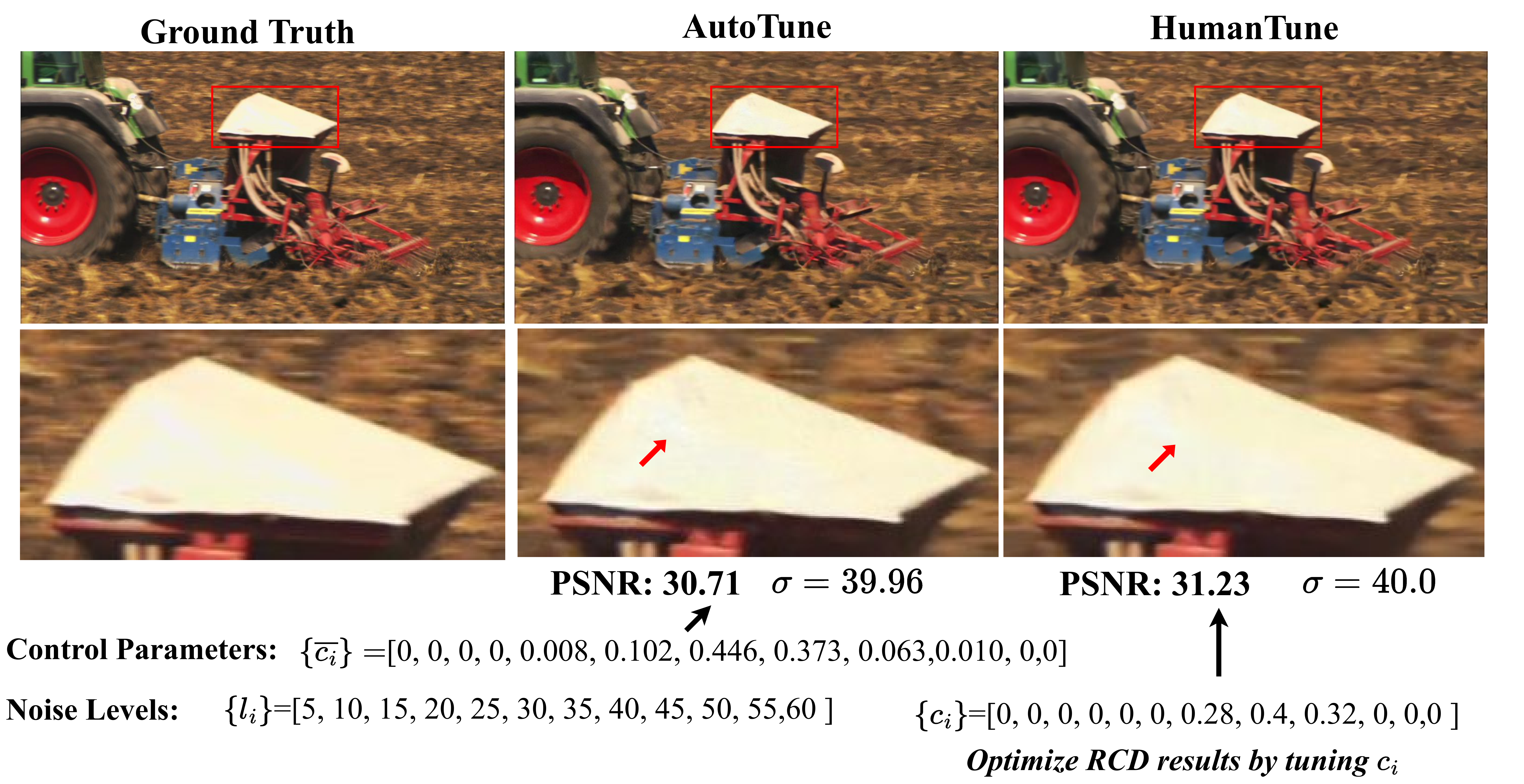}
\centering
\caption{\small Example of RCD denoising results by AutoTune and HumanTune on Set8. AutoTune module provides reference control parameters, \ie, $\{ \bar{c}_i \}_{i=1}^L$), to generate the denoising result, and it can be further improved by fine-grained artificial tuning  (HumanTune), \ie, $\{c_i\}$, without changing the noise intensity  (both $\sigma=40$).}
\label{fig:AutoTune}
\vspace{-10pt}
\end{figure}

\subsection{New Cardinality for Denoising Control. }
\label{sec:3.6}
Unlike existing methods that only modulate noise intensity, our RCD control scheme allows users to further optimize the denoising result to a given noise intensity by tuning $\{c_i\}$, as long as the weighted mean of $l_i$ towards $c_i$ remains the same.
Eqn.~\ref{eq:var} shows that when $ \sum_{i=1}^L  c_i^2 \mathbf{Var} (\Tilde{\mathcal{N}}_i) = \sum_{i=1}^L  c_i^2 l_i^2  $ is fixed, the variance of the output noise $\mathbf{Var}  (\sum_{i=1}^{L} c_i \Tilde{\mathcal{N}}_i )$ would be also fixed. 
By exploring $c_i$ under the condition of fixed $\sum_{i=1}^L  c_i^2 l_i^2 $, we can further optimize the denoising results at a specific noise level by 
involving different components  ($\mathcal{N}_i$) in the noise interpolation. 
As shown in Fig.~\ref{fig:AutoTune}, our AutoTune module can generate high-quality results using just the reference control parameters $\{ \bar{c_i} \}$. Users can further improve the result  by artificially tuning $\{c_i\}$ around $\{ \bar{c_i} \}$, even at the same noise level. 
%

%

\begin{table*}[h!]
\small 
  \begin{center}
  \vspace{-15pt}
    \caption{\small Gaussian single image denoising results  (PSNR). RCD is evaluated with AutoTune results. ``-'': not reported}
    \label{tab:image}
    \begin{tabular}{c|c|c|c|c|c} 
    \hline
      \multirow{2}{*}{\textbf{Method}} & \multirow{2}{*}{\textbf{Controllable}} & \textbf{CBSD68} & \textbf{Kodak24} & \textbf{McMaster} & \textbf{Urban100} \\
       & &$\sigma$ = 15 $\sigma$ = 25 $\sigma$ = 50 & $\sigma$ = 15 $\sigma$ = 25 $\sigma$ = 50 & $\sigma$ = 15 $\sigma$ = 25 $\sigma$ = 50 & $\sigma$ = 15 $\sigma$ = 25 $\sigma$ = 50\\
      \hline
      IRCNN~\cite{zhang2017learning} & \xmark   &33.86  ~  31.16  ~  27.86 &  34.69   ~  32.18  ~   28.93  & 34.58  ~  32.18   ~  28.91 &  33.78  ~  31.20  ~   27.70 \\
      FFDNet~\cite{zhang2018ffdnet} &\xmark  & 33.87   ~  31.21  ~  27.96  & 34.63   ~ 32.13  ~   28.98 &  34.66   ~  32.35  ~   29.18  & 33.83    ~  31.40  ~   28.05 \\
      DnCNN~\cite{zhang2017beyond}  & \xmark  &33.90   ~  31.24  ~   27.95 &  34.60   ~  32.14  ~   28.95 &  33.45   ~  31.52  ~   28.62 &  32.98  ~  30.81   ~  27.59 \\
      DSNet~\cite{peng2019dilated}  &\xmark   &33.91    ~  31.28  ~   28.05 &  34.63  ~  32.16 ~    29.05  & 34.67   ~  32.40  ~   29.28 &  ~~~~-~~~~  \\
      CResMD~\cite{he2020interactive} &\checkmark & 33.97   ~ ~~~~-~~~~ ~  28.06 & ~~~~-~~~~ & ~~~~-~~~~ & ~~~~-~~~~ \\
      AdaFM-Net~\cite{he2019modulating} & \checkmark & 34.10 ~ 31.43 ~ 28.13 &~~~~-~~~~ & ~~~~-~~~~ & ~~~~-~~~~\\ \hline
      NAFNet~\cite{chen2022simple} & \xmark &34.11 ~ 31.49 ~ 28.27 & 35.14 ~ 32.70  ~  29.68 & 35.07 ~  32.82 ~  29.79   & 34.41  ~  32.09  ~  29.00   \\
      \textbf{NAFNet-RCD (ours)} & \checkmark &\textbf{34.13  ~  31.49  ~  28.26} & \textbf{35.15  ~  32.72  ~  29.69} & \textbf{35.11  ~  32.84  ~  29.81} &  \textbf{34.45 ~  32.12  ~  29.02}\ \\ \hline
    \end{tabular}
  \end{center}
  \vspace{-15pt}

\end{table*}


\subsection{Optimization Targets}
To guarantee the visual quality of arbitrarily edited noise $\sum_{i=1}^{L} c_i \Tilde{\mathcal{N}}_i$ for any given set of control parameters $\{ c_i \}_{i=1}^L$, we adopt a multi-level concurrent training strategy by minimizing the difference between each level's noise output and the ground truth noise, \ie, $\mathcal{L}_{level}$, which is derived by:
\begin{equation}
    \mathcal{L}_{level} =  \frac{1}{L}\sum^L_{i=1} \mathcal{L} (\mathbf{I}_{gt}, \mathbf{I_n} +  \Tilde{\mathcal{N}}_i ) 
\end{equation}
where $\mathbf{I}_{gt}$, $\mathbf{I_n}$ are ground truth clean image and the input noisy image. $\mathcal{L} (.)$ can be any loss functions (e.g., L2 loss or PSNR loss).  

Spectacularly, $\mathbf{I_n} + \Tilde{\mathcal{N}}_i$ can be regarded as corner cases of RCD when we use one-hot control parameters as input. Joint optimization of all the noise levels ensures that each element of $\mathcal{N}_i$ can be trained as optimal noise estimation under the condition of fixed noise level $l_i$.

Together with the AutoTune module optimization, our final cost function can be written as:
\begin{equation}
    \mathcal{L}_{total} =   \lambda \mathcal{L}_{level} +   \mathcal{L} (\mathbf{I}_{gt}, \mathbf{I_n} +  \sum_{i=1}^{L} \bar{c}_i \Tilde{\mathcal{N}}_i ),
\end{equation}
where $\lambda$ is the loss weight  ($\lambda = 0.1$ in our experiments), and $\mathcal{L} (\mathbf{I}_{gt}, \mathbf{I_n} +  \sum_{i=1}^{n} \bar{c}_i \Tilde{\mathcal{N}}_i ) $ optimizes the denoising result derived by the model-suggested control parameters.

\section{Experiments}

This section is organized as follows: First, we demonstrate the effectiveness of our plug-in RCD with SOTA image denoising methods~\cite{chen2022simple} in different scales on synthetic noise datasets. Next, to evaluate the ability of blind denoising on real-world data, we conduct experiments on popular real-world denoising dataset SIDD~\cite{abdelhamed2018high}. Then, we apply our real-time controllable RCD pipeline on video denoising applications. At last, we empirically discuss some design details described in the previous sections.

\begin{figure*}[h]
\vspace{-15pt}
\includegraphics[width=1\linewidth]{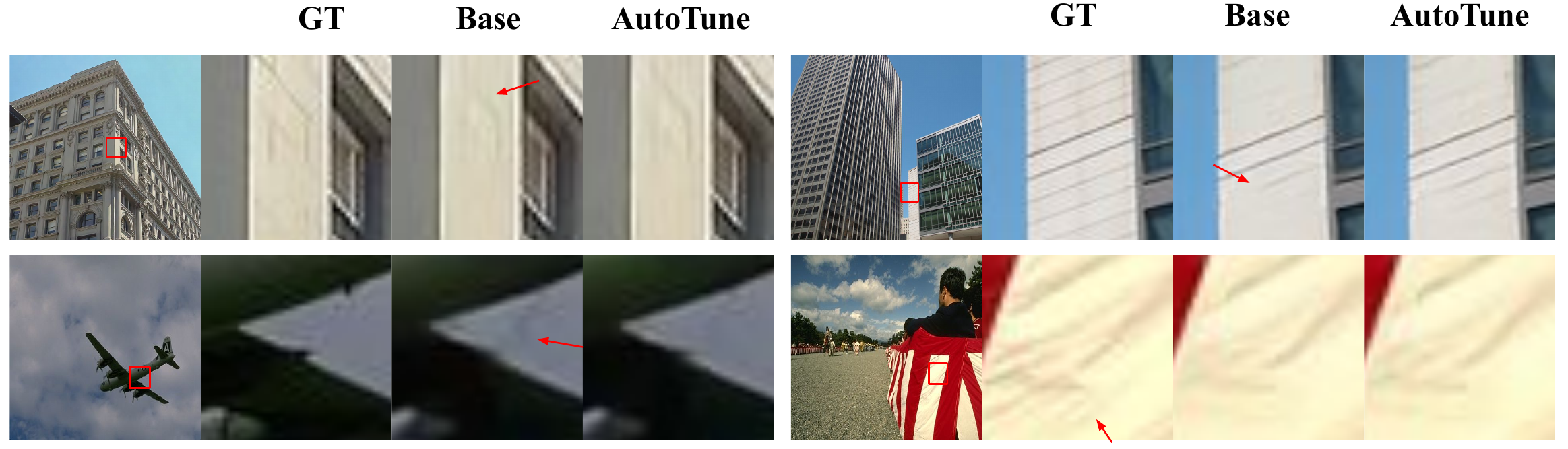}
\centering
\caption{\small Visual comparison of RCD and their baseline results on $\sigma=50$ denoising. \textbf{GT}: Ground truth.  \textbf{Base}: Baseline model without RCD. \textbf{AutoTune}: RCD results by applying control parameters from AutoTune module.}
\label{fig:results}
\end{figure*}

\subsection{Gaussian Single Image Denoising}


\label{sec:4.1}
\textbf{Experimental Setup.}
To fully demonstrate the effectiveness of the proposed RCD, we choose the most recent SOTA method NAFNet~\cite{chen2022simple} as our backbone. 
Following~\cite{zamir2022restormer}, we first conduct denoising experiments on several widely-used synthetic color image benchmarks  (DIV2K~\cite{agustsson2017ntire}, BSD400~\cite{MartinFTM01}, Flickr2K~\cite{young2014image} and WaterlooED~\cite{ma2016waterloo}) with additive white Gaussian noise ($\sigma \in [0, 60]$). The training patch size is 128 × 128 and the batch size is 64. 
We train our model with Adam~\cite{kingma2014adam} optimizer and  learning rate $1e-3$  for total 60K
iterations.
Consistent to~\cite{chen2022simple}, PSNR loss is adapted as the loss function. 
Both the baseline model  (NAFNet) and its RCD variants  (NAFNet-RCD) are trained from scratch.
For settings of RCD, we initialize $L=12$ and $\{ l_i \} = [5, 10, ..., 60]$ for synthetic denoising training.

\begin{table}[]
  \small 
  \centering
    \caption{\small Running time comparison for RCD and other controllable methods during test time. \textbf{Full Pipeline} compares full pipeline latency for the model to infer 1000 images, and \textbf{Edit-only} compares latency for editing one image with 1000 different control parameters.   }
    \label{tab:time}
    \scalebox{0.9}{
    \begin{tabular}{c | c |c|c} 
    \hline
       \textbf{Method} &\textbf{Multi-stage training} &\textbf{Full Pipeline} & \textbf{Edit-only}\\
      \hline
      AdaFM-Net & required  & 81.03s & 81.03s\\
      CResMDNet & not required & 128.08 & 128.08s\\
      \textbf{NAFNet-RCD}  &  not required & \textbf{64.84s} & \textbf{0.04s}\\
      \hline
    \end{tabular}
    }
    \vspace{-12pt}
\end{table}

\begin{table*}
  \begin{center}
    \caption{\small Ablation of RCD on various backbone sizes. }
    \label{tab:size}
    \begin{tabular}{c|c|c|c|c} 
    \hline
      \multirow{2}{*}{\textbf{Method}} & \textbf{CBSD68} & \textbf{Kodak24} & \textbf{McMaster} & \textbf{Urban100} \\
       & $\sigma$ = 15 $\sigma$ = 25 $\sigma$ = 50 & $\sigma$ = 15 $\sigma$ = 25 $\sigma$ = 50 & $\sigma$ = 15 $\sigma$ = 25 $\sigma$ = 50 &
       $\sigma$ = 15 $\sigma$ = 25 $\sigma$ = 50\\
      \hline
      NAFNet-tiny & 33.58 30.91 27.62 & 34.33 31.84 28.63 &  33.85 31.61 28.55 & 32.96 30.37 26.92\\
      \textbf{ NAFNet-RCD-tiny} & \textbf{33.71} \textbf{31.06} \textbf{27.68} & \textbf{34.46 31.98 28.65}  &  \textbf{34.07 31.78 28.61} &\textbf{ 33.22 30.66 27.18}\\ \hline
      NAFNet-small & 33.84 31.18 27.91 & 34.68 32.18 29.01 & 34.68 32.18 29.01 & 33.61 31.10 27.68\\
       \textbf{NAFNet-RCD-small} & \textbf{33.96} \textbf{31.31} \textbf{28.05} & \textbf{34.83 32.32 29.14} & \textbf{34.71 32.40 29.26} &  \textbf{33.92 31.46 28.08}\\ \hline
      NAFNet & 34.11 31.49 28.27 & 35.14 32.70 29.68 & 35.07 32.82 29.79   & 34.41 32.09 29.00 \\
       \textbf{NAFNet-RCD }& \textbf{34.13 31.49 28.26} & \textbf{35.15 32.72 29.69} & \textbf{35.11 32.84 29.81} &  \textbf{34.45 32.12 29.02}\\ \hline
    \end{tabular}
    \vspace{-15pt}
  \end{center}
\end{table*}

\textbf{Complexity analysis.}
%
Extensive adjustments of controllable parameters are often required to obtain one satisfying result for users. 
Therefore, editing time is vital for controllable methods.  
This section compares the inference and editing latency of our RCD  and conventional controllable pipelines on GTX 1080Ti. 
As shown in Tab.~\ref{tab:time}, the proposed RCD  not only outperforms  other conventional controllable pipelines on inference time, but more importantly, can overwhelm those traditional controllable designs on editing time, which can be more than \textbf{2000} times faster  (as editing process of RCD is network-free, without reliance on sub-networks). This comparison confirms that our RCD is more than enough for real-time image editing.

\textbf{Results Analysis.} 
We evaluate our proposed method on widely used synthetic noise datasets CBSD68~\cite{martin2001database}, Kodak24~\cite{franzen1999kodak}, McMaster~\cite{zhang2011color} and Urban100~\cite{huang2015single}  with noise levels $\sigma (15)$, $\sigma (15)$ and $\sigma (50)$. 
 RCD is evaluated with denoising results using AutoTune outputs$\{ \bar{c}_i\}$.
As shown in Tab.~\ref{tab:image}, NFANet-RCD achieves comparable performance to the baseline NFANet consistently on multiple datasets, 
indicating that our plug-in RCD module enables real-time controllable denoising for NAFNet without sacrificing its original denoising performance. 
%
%
Please note that NAFNet-RCD can yield comparable results to the backbone just by using the AutoTune outputs, and
the performance can be further improved by manually tuning the control parameters  ( See Sec.~\ref{sec:3.6}.)
 We further show the qualitative performance of NAFNet-RCD in Fig.~\ref{fig:results}. 
 NAFNet-RCD can recover more details of some degraded images, 
 which may be benefited from RCD's richer representation capacity by integrating multiple noise maps. 

\textbf{Slimmer Model Variants.}
Towards the goal of evaluating the compatibility and robustness of RCD, we conduct ablations by applying RCD to different-sized backbones. 
Specifically, we shrink the width and block numbers of NAFNet, denoting derived models as NAFNet-small ($\frac{1}{4} \times$) and NAFNet-tiny ($\frac{1}{16} \times$).
Tab.~\ref{tab:size} reports the results of RCD with those scaled backbones.
It can be observed that the RCD-variants can achieve comparable and even slightly better denoising results compared to their baselines, which further demonstrates RCD's robustness and effectiveness for different-sized backbones.
%


\begin{table}[h!]
  \begin{center}
    \caption{\small Image denoising results on SIDD. \textbf{Real noise}: results on real-world SIDD test sets. \textbf{Synthetic noise}: results on SIDD test set with additive Gaussian noise  ($\sigma=25$). }
    \label{tab:real}
    \begin{tabular}{c|c c|c c} 
    \hline
      \multirow{2}{*}{\textbf{Method}} & \multicolumn{2}{c|}{\textbf{Real noise}} & \multicolumn{2}{c}{\textbf{Synthetic noise}}  \\
      & \textbf{PSNR} & \textbf{SSIM} & \textbf{PSNR} & \textbf{SSIM} \\
      \hline
      NAFNet-tiny & \textbf{42.19} & \textbf{0.9796} & 38.46  & 0.9551  \\
      \textbf{NAFNet-RCD-tiny} & 41.86 & 0.9781 & \textbf{38.60 }& \textbf{0.9558}  \\ \hline
      NAFNet & \textbf{43.22} & \textbf{0.9818}  & 38.85 & 0.9481 \\
      \textbf{NAFNet-RCD} & 42.91 & 0.9806 & \textbf{39.14} & \textbf{0.9580 } \\
      \hline
    \end{tabular}
      \vspace{-15pt}
  \end{center}
\end{table}

\subsection{Real Single Image Denoising}
\textbf{Experimental Setup (Real Image) }
Unlike existing controllable denoising methods~\cite{he2019modulating, wang2019cfsnet}  which focus on synthetic benchmarks, we are the first solution that attempts to extend controllable denoising to real-world SIDD datasets.
SIDD consists of real noisy images captured by smartphones with $\sigma \in [0, 50]$.
Instead of using full SIDD data, we choose subsets of SIDD with $\sigma \in [0, 12]$  (around 70\% of the entire dataset) to train our RCD model, which is initialized with $L=4$ and $\{ l_i\} = [3, 6, 9, 12]$.
The main reason is the lacking of high $\sigma$ data at given levels in SIDD because of SIDD's highly long-tailed noise level distribution.
Specifically, most noisy images in SIDD gather in $\sigma<12$ and the samples distribute sparsely when $\sigma$ is large. 
Consistent to Sec.~\ref{sec:4.1}, we adopt NAFNet  (SOTA methods for SIDD challenge~\cite{chen2022simple}) as our backbone at two scales  ($1\times$, $\frac{1}{16}\times$).
Both NAFNet-RCD and the corresponding baselines are trained on this subset with the same training settings as in~\cite{chen2022simple}.


%
%

%

%
%
%

\begin{figure*}[h!]
\vspace{-20pt}
\includegraphics[width=1\linewidth]{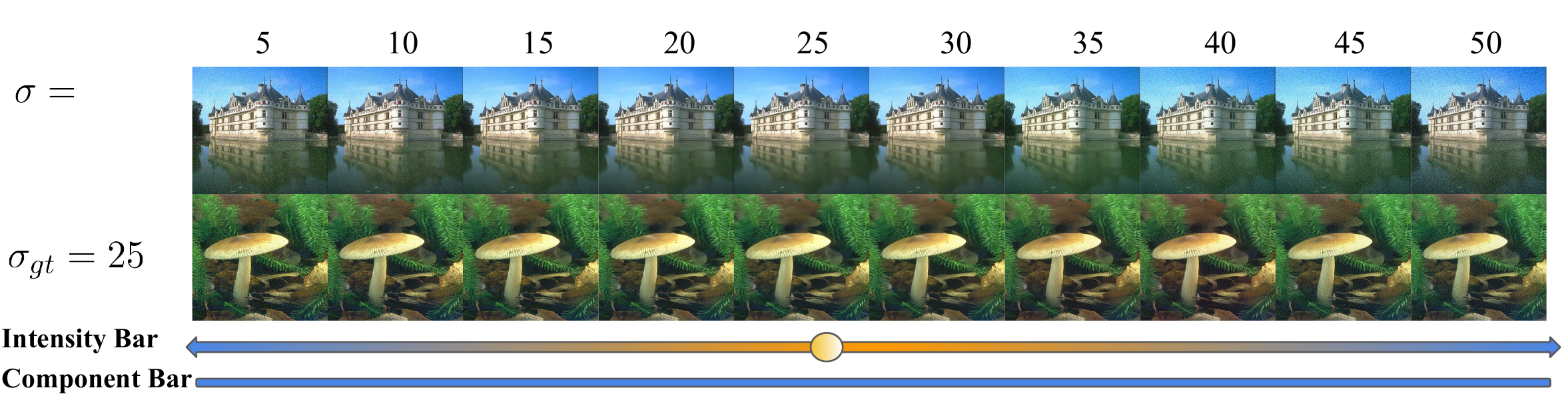}
\centering
\caption{\small Illustration of RCD control logics. Users can retouch the \textbf{denoising level} by tuning the \textbf{Intensity} bar ($\sigma = \sqrt{\sum_{i=1}^L  c_i^2 l_i^2 }$ ) and setup their perceptual preference at \textbf{fixed level} by tuning \textbf{Component} bar  (changing $\{c_i\}$ while keeping $\sigma$).   }
\label{fig:level}
  \vspace{-10pt}
\end{figure*}

\textbf{Results and Analysis}
We conduct blind denoising experiments on SIDD with different RCD model scales to evaluate its adjustability to the real-world dataset.
As shown in Tab.~\ref{tab:real}  (left), our RCD  (AutoTune results) can achieve high-quality controllable real-world denoising in both model scales.
However, we note that enabling controllable denoising with RCD may still result in a slight decrease in quantitative results  (about 0.3dB), which may  be a result of unbalanced data for each level and short noise level interval ($|l_{i+1} - l_i|$, see more discussion in Sec.~\ref{sec:4.4}).

\textbf{SIDD with synthetic noise.}
We extensively conduct synthetic denoising experiments on SIDD to further show the compatibility of RCD on SIDD datasets.
Following Sec.~\ref{sec:4.1}, we add random Gaussian noise $\sigma \in [0, 60]$ to SIDD training data, and both methods are evaluated on $\sigma=50$ SIDD test samples. 
As shown in Tab.~\ref{tab:real}  (right), RCD models slightly outperform their baselines, demonstrating RCD's compatibility for SIDD. 
Moreover, this result can also indicate that RCD's performance drop on SIDD real image may arise from the noise distribution and RCD configurations, rather than RCD's adaptive capacity to SIDD data. %
See Appendix for more results and visualizations.

\subsection{Video Denoising}

\textbf{Experiment Setup}
Following common practice~\cite{tassano2020fastdvdnet,sheth2021unsupervised, liang2022vrt}, we train our models on DAVIS training set and use DAVIS-test and Set-8 for benchmarking.
Like in~\cite{tassano2020fastdvdnet}, we add Gaussian noise with random standard deviation between 5-50 on the DAVIS clean videos for training. 
The DAVIS set contains 30 color sequences of resolution $854 \times 480$, which will be randomly cropped into $128\times128$ patches during training. 
%
%
Other training settings and hyperparameters are kept the same as~\cite{tassano2020fastdvdnet} for a fair comparison.

\textbf{Choice of Basic model.} 
We choose FastDVD~\cite{tassano2020fastdvdnet} as our backbone model. Although recent methods~\cite{vaksman2021patch, liang2022vrt} outperform FastDVD by 1-2 PSNR  at most, they actually introduce huge models and extra heavy operations like patch clustering~\cite{vaksman2021patch} and layer-wise frame-to-frame wrapping using optical flow~\cite{liang2022vrt}  (  $>100\times$ slower than FastDVD).

\textbf{Results and Analysis.}
Like~\cite{sheth2021unsupervised}, we evaluate our video denoising models with the input length of one frame and five frames.
We denote RCD models for video denoising as ``FastDVD-RC'' and compare their quantitative AutoTune denoising results to baseline FastDVD in Tab.~\ref{tab:video}. 
Consistent with preceding sections, AutoTune results of FastDVD-RCD can demonstrate comparable performance to the default FastDVD, which means our RCD can also achieve lossless real-time noise editing in video scenarios.
Unlike previous heavy controllable denoising methods, our real-time RCD can even allow users to do online video denoising editing without any latency.

\begin{table}[h!]
  \small 
  \centering
  \vspace{-5pt}
    \caption{\small Video denoising results.  }
    \label{tab:video}
    \scalebox{0.8}{
    \begin{tabular}{c|c|c  c |c c} 
    \hline
       \multirow{2}{*}{Test set}  & \multirow{2}{*}{ $\sigma$}  & \multicolumn{2}{c|}{1 frame}  & \multicolumn{2}{c}{5 frames}  \\
      
      & &  FastDVD & FastDVD-RCD &  FastDVD  & FastDVD-RCD\\      \hline

         \multirow{4}{*}{DAVIS}  & 20 & 34.17 & \textbf{34.21} & \textbf{35.69} & 35.65 \\
             & 30 & 32.45 &  \textbf{32.69} & \textbf{34.06} & 34.04  \\
             & 40 & 31.39 &  \textbf{31.60} & \textbf{32.80} & 32.78  \\
             & 50 & 30.26 &  \textbf{30.57}  & 31.83 & \textbf{31.85}  \\
        \hline
         \multirow{4}{*}{Set 8} & 20 & 31.99 & \textbf{32.01}  & 33.43 &  \textbf{33.46} \\
         & 30 &  30.61 & \textbf{30.65}  & 31.62 & \textbf{31.71} \\
        & 40 & 29.62 & \textbf{29.83}  & 30.36 & \textbf{30.42} \\
         & 50 & 28.61 & \textbf{28.85}   & 29.41 & \textbf{29.60}  \\
         \hline
    
    \end{tabular}
    }
    \vspace{-10pt}
\end{table}
%

%

\subsection{Discussions}

\textbf{Selection of Denoising Levels.}
\label{sec:4.4}
Differing from conventional denoising methods, RCD requires  a group of predefined noise levels $\{l_i\}_{i=1}^L$. 
To evaluate how the selection of $\{l_i\}_{i=1}^L$ affects RCD's performance, we conduct ablation studies on FastDVD-RCD by changing the number of noise maps  ($L$)  (See Tab.~\ref{tab:level}.)
All of the models are trained on noisy images with $\sigma \in  (0, 60]$ and uniformly sampled noise levels that $\{l_i = \frac{60}{L} * i \}_{i=1}^L$.
We observe that larger $L$ means more fine-grained control on denoising, but it may incur a performance drop. 
In fact, when $n$ is large we find that $\mathcal{L}_{level}$ will also keep large and be hard to optimize.
Trading-off performance and control precision, we empirically choose $L=12$ and noise level interval $|l_{i+1} - l_{i}| = 5$ as defaults.


\begin{table}[h!]
  \small 
  \vspace{-5pt}
  \centering
    \caption{\small Ablations of FastDVD-RCD on different number of noise levels. Reported scores are   PSNR of AutoTune outputs and GT.}
    \label{tab:level}
    \scalebox{0.9}{
    \begin{tabular}{c|c|c  c c c c} 
      \hline
      Test Set & $\sigma$ & $L=1$ & $L=2$ & $L=12$ & $L=30$ & $L=60$ \\ \hline
      \multirow{4}{*}{Set8} & 20 & 31.87 & 31.42 & \textbf{32.01} & 31.39 & 31.07 \\
       & 30 & 30.51 & 30.09 & \textbf{30.65} & 30.12 & 29.75 \\
       & 40 & 29.60 & 29.31 & \textbf{29.83} & 29.33 & 29.01 \\
       & 50 & 28.62 & 28.29 & \textbf{28.85} & 28.22 & 28.05 \\ \hline
    \end{tabular}
    }
    \vspace{-5pt}
\end{table}

\textbf{Control Capacity.}
This section discusses the representation capacity of $c_i$ as control parameters. 
Generally, $c_i$ controls the denoising process on two aspects: intensity and components.
Firstly, the noise levels of RCD outputs are identical and can be derived by $\sigma = \sqrt{\sum_{i=1}^L  c_i^2 l_i^2 }$  (See Sec.~\ref{sec:3.4}), which allows us to control the denoising intensity by changing $\{c_i\}$.
Fig.~\ref{fig:level} depicts visualizations of RCD-controlled denoising under different intensity settings.
Besides, as discussed in Sec.~\ref{sec:3.6}, RCD supports further  optimization of the denoising results at specific noise intensity by tuning $c_i$ by involving different  components of $\Tilde{\mathcal{N}_i}$.  (Please be reminded that $\Tilde{\mathcal{N}_i}$ is trained by $\mathcal{L}_{level}$, denoting learned optimal denoising results at every fixed level $l_i$.)



\section{Summary}
We present RCD framework that enables real-time noise editing for controllable denoising.
Unlike existing continual-level denoising methods, RCD doesn't require multiple training stages and auxiliary networks. 
With the proposed Noise Decorrelation module, RCD transforms the control of denoising into white-box operations, with no requirement to feed control parameters to networks at test time, which enables real-time editing even for heavy network models.
Extensive experiments on widely-used real/synthetic image and video denoising datasets demonstrate the robustness and effectiveness of our RCD.
 
 \section{Acknowledgement}
This paper is partially supported by the National Key R\&D Program of China No.2022ZD0161000,  the General Research Fund of HK No.17200622, and Shanghai Committee of Science and Technology  (Grant No. 21DZ1100100).

\clearpage
{\small
\bibliographystyle{ieee_fullname}
\bibliography{egbib}
}

\clearpage

\newpage

\appendix

\twocolumn[{
\begin{figure}[H]
\setlength{\linewidth}{\textwidth}
\setlength{\hsize}{\textwidth}
\includegraphics[width=\linewidth]{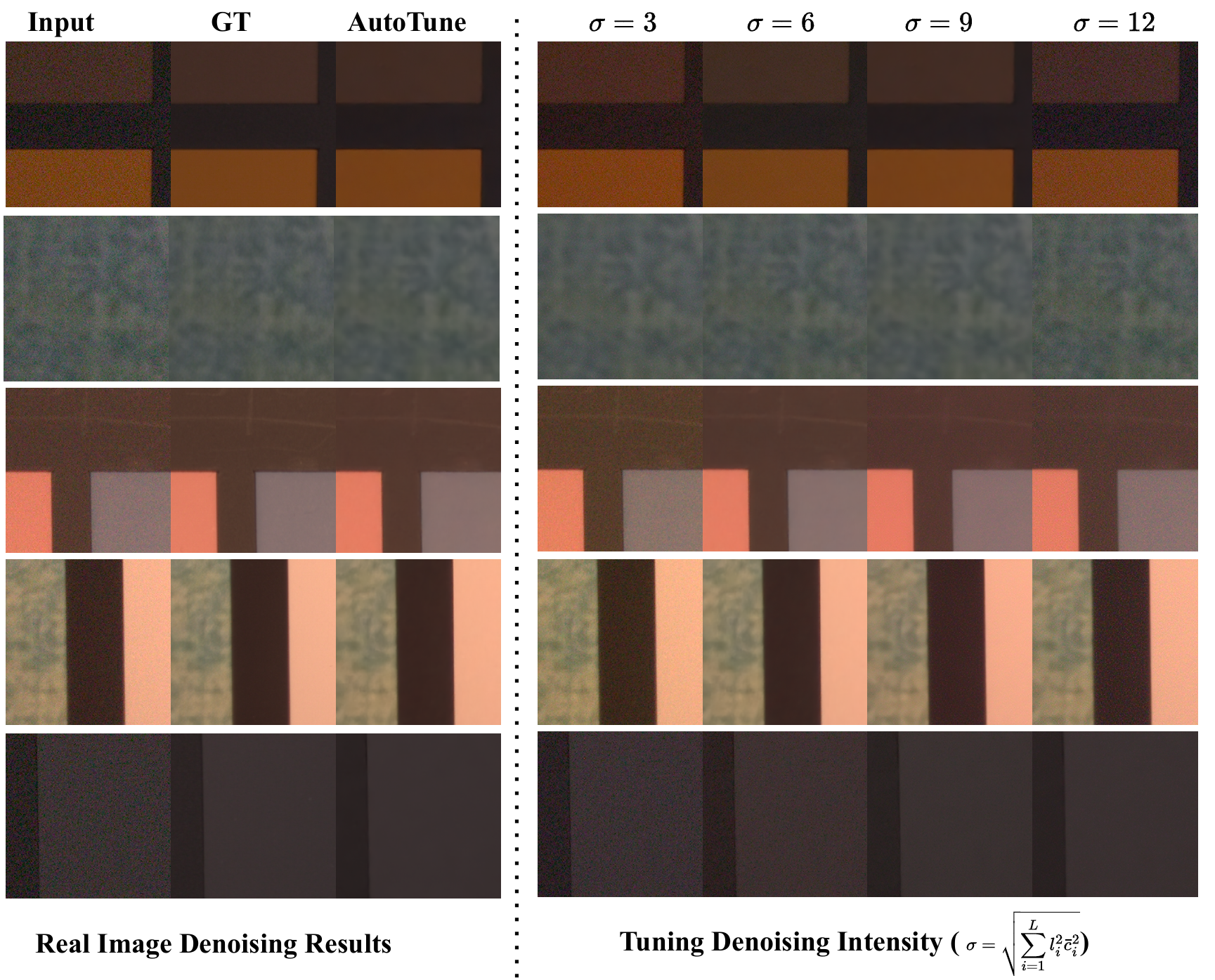}
\centering
\caption{\small RCD real image denoising results on SIDD. GT: ground truth. AutoTune: AutoTune results of RCD.}
\label{fig:real}
\end{figure} 
}]

\begin{figure*}[ht!]
\includegraphics[width=1\linewidth]{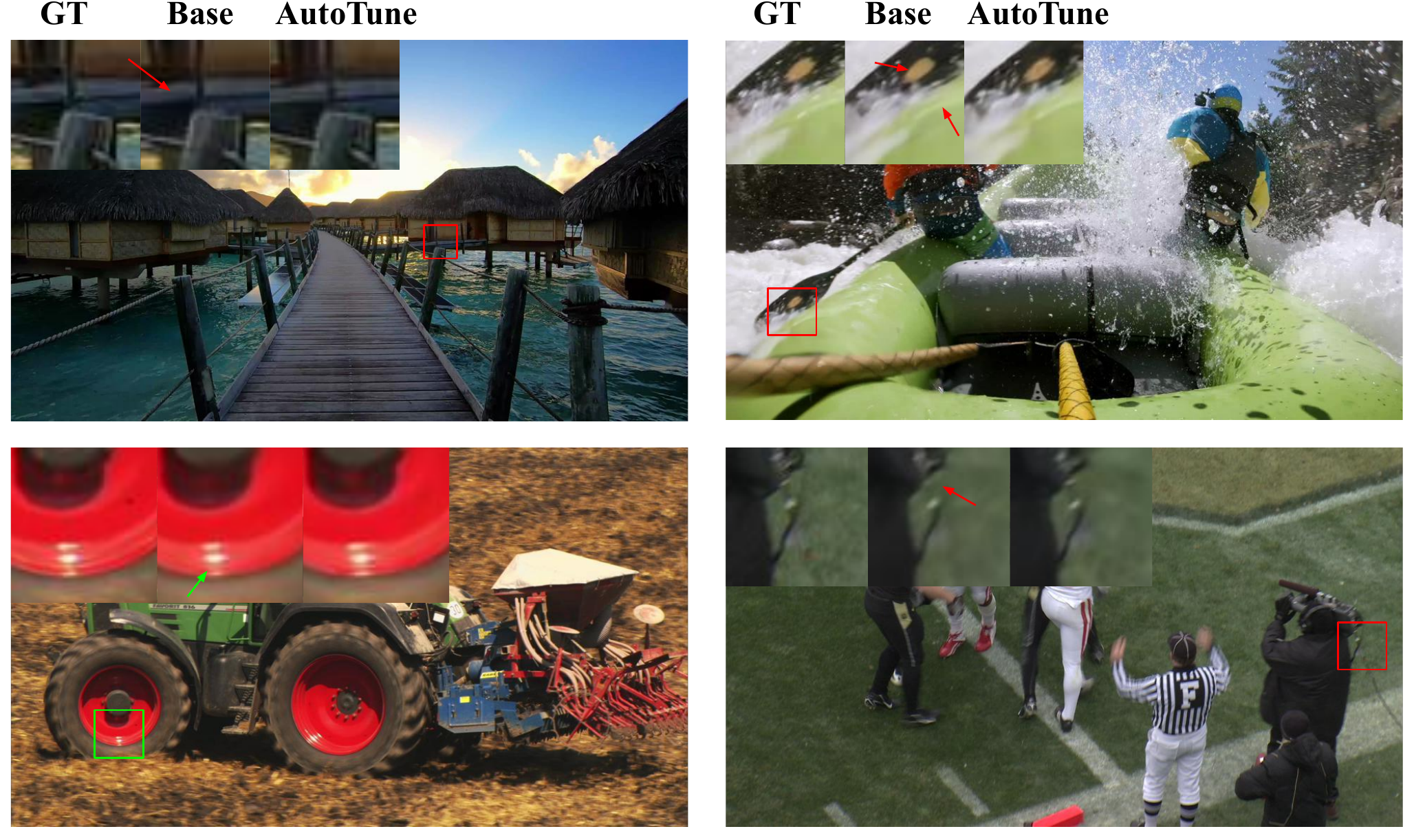}
\centering
\caption{\small Video denoising results. Base: uncontrollable baseline.   }
\label{fig:video}
\end{figure*}

\section{More Visualization Results}
This section shows more visual results to demonstrate the effectiveness of our proposed RCD.  Besides, we also provide \textbf{demo video} for showing features of RCD (Please see in attached files of supplementary material).

\textbf{Real Image Denoising.}
We visualize RCD denoising results on SIDD test set in Fig.\ref{fig:real}.
The left part shows the comparison  of RCD (AutoTune) and real image ground truth,  and the right part gives RCD results by tuning the noise level. 
As demonstrated, RCD can support controllable real image denoising and yield high visual quality results.


\textbf{Video denoising results.}
We further show the qualitative performance of FastDVD-RCD in Fig.\ref{fig:video}, with comparison to baseline uncontrollable FastDVDnet.
Consistent with image denoising, FastDVD-RCD  can recover more details of some degraded images, which may be benefited from RCD’s richer representation capacity by integrating multiple noise maps

\begin{figure}[ht!]
\includegraphics[width=1\linewidth]{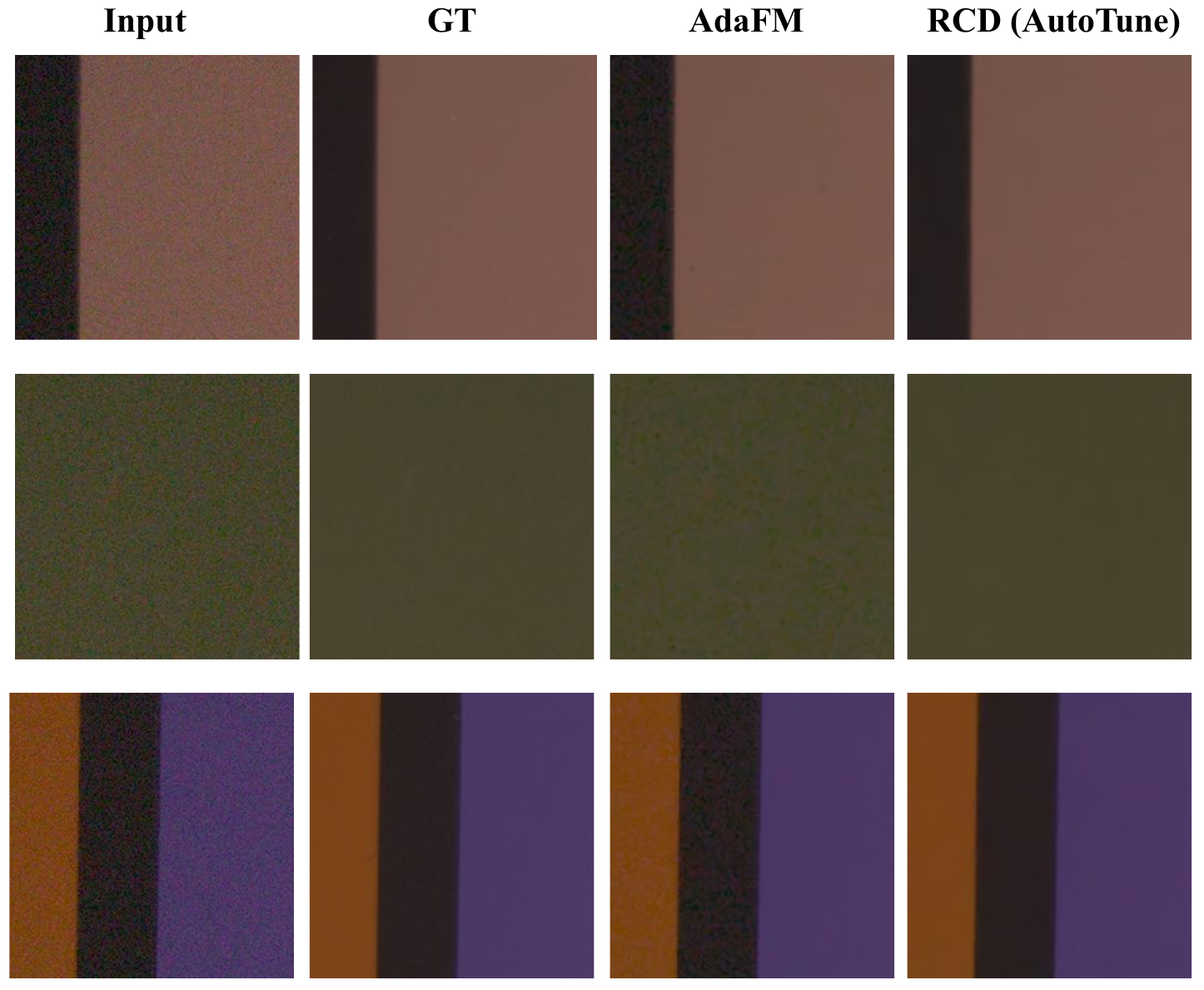}
\centering
\caption{\small Comparison of RCD and AdaFM on SIDD real image denoising. }
\label{fig:adafm}
\vspace{-10pt}
\end{figure}

\textbf{Comparison of RCD and AdaFM on SIDD.}
In Fig.\ref{fig:adafm} we show the comparison of RCD and representative conventional controllable denoising method AdaFM. 
Compared to RCD and GT, AdaFM results have more artifacts and remained noises.

\section{Implementation Details}

\textbf{Choice of Basic Image Model}. 
Considering the application of real-time image controllable denoising, we need to select the base models with acceptable running time and parameters. Besides, the base model is required to support prior-free blind denoising to be applied on real-world data and should be able to be trained in an end-to-end manner for level loss to be pluged-in readily. 
Our models on single image denoising are based on  the SOTA restoration model NAFNet\cite{chen2022simple}, which can be scaled flexibly from 1.1 GMACs to 65 GMACs.
To balance the running time and the performance, we adopt NAFNet with width 16/32 and number of blocks 8. We also conduct experiments on NAFNet base model with width 32 and number of blocks 36 to verify the effectiveness of RCD on relatively larger model.

\textbf{Alteration of Basic Model}
Only two adjustments of the base model are required to support our proposed editable denoising.
First, we alter the output channel number of the base model ending layer from output image channel number to  $L \times$ output image channel number, $L$ is the number of the predefined noise level. For example, when training noise level from 0 to 60, the uniform noise level gap between each noise map is 5 and there are 13 of them in total ([0,5,10,...55,60]). Then we apply noise decorrelation on those fixed-level noise maps generated by our model and also feed them to the AutoTune module.
Second, for your AutoTune module, we add an additional CNN layer to predict a feature-map. After the adaptive average pooling layer and temperatured softmax activation, we attain a series of weights to fuse the 13 noise maps as model-suggested guidance to the user.

\textbf{Model Variants Details.}
We calculate the parameters of NAFNet base model and our NAFNet-RCD model. Compared with NAFNet, NAFNet-RCD only has alterations on two CNN layers, which is negligible for normal-size networks. Specifically, these additional parameters account for 0.03\%, 3.60\%, and 9.17\% of total parameters for model size 1) NAFNet-RCD: width 32, number of blocks 36, 2) NAFNet-RCD-small: width 32, number of blocks 8, 3) NAFNet-RCD-tiny: width 16, number of blocks 8. Please be noted that the number of additional parameters is only 7.7K even though it accounts for 9.17\% of light model NAFNet-RCD-tiny. 

\section{Additional Experiments}

\textbf{ Results on other real-world datasets.}
We further evaluate our RCD models on the PolyU~\cite{xu2018real} and Nam~\cite{nam2016holistic} benchmarks. Both the RCD and baseline models are trained on SIDD real-world data. Table~\ref{tab:real-more} shows that on both benchmarks, the RCD models can still perform controllable denoising without sacrificing much performance, and on Nam, the RCD models even slightly outperform their uncontrollable baselines.

\begin{table}[h!]
  \begin{center}
    \caption{\small Image denoising results on PolyU and Nam. \textbf{PolyU}: results on real-world PolyU test sets. \textbf{Nam}: results on real-world Nam test set. }
    \vspace{-8pt}
    \label{tab:real-more}
    \scalebox{0.8}{
    \begin{tabular}{c|c c|c c} 
    \hline
      \multirow{2}{*}{\textbf{Method}} & \multicolumn{2}{c|}{\textbf{PolyU}} & \multicolumn{2}{c}{\textbf{Nam}}  \\
      & \textbf{PSNR} & \textbf{SSIM} & \textbf{PSNR} & \textbf{SSIM} \\
      \hline
      NAFNet-tiny & 38.52 & \textbf{0.9827} & 38.93  & 0.9881  \\
      \textbf{NAFNet-RCD-tiny} & 38.36 & 0.9826 & \textbf{39.03 }& 0.9881  \\ \hline
      NAFNet & 39.11 & 0.9837  & 39.54 & 0.9894 \\
      \textbf{NAFNet-RCD} & 39.07 & 0.9837 & \textbf{39.67} & \textbf{0.9896 } \\
      \hline
    \end{tabular} }
  \end{center}
\end{table}

\textbf{Results on more architectures.}
We also test our RCD using the Restormer~\cite{zamir2022restormer} model. As shown in Table~\ref{tab:resformer}, Restormer-RCD slightly outperforms its baseline, consistent with NAFNet.

\begin{table}[h!]
  \begin{center}
    \caption{\small Restormer results on SIDD test set with additive Gaussian noise ($\sigma$ from 0 to 50 ).}
    \label{tab:resformer}
                  \vspace{-10pt}
    \scalebox{0.9}{

              \vspace{-10pt}    \begin{tabular}{c|c c} 
    \hline
      \multirow{2}{*}{\textbf{Method}} & \multicolumn{2}{c}{\textbf{SIDD Synthetic noise }} \\
      & \textbf{PSNR} & \textbf{SSIM} \\
      \hline
      Restormer & 41.31 & 0.9763  \\
      \textbf{Restormer-RCD} & \textbf{41.79} & \textbf{0.9781}   \\ \hline
    \end{tabular}
    }
  \end{center}
\end{table}

\end{document}